\def\year{2022}\relax
\documentclass[letterpaper]{article} 
\usepackage{aaai22}  
\usepackage{times}  
\usepackage{helvet}  
\usepackage{courier}  
\usepackage[hyphens]{url}  
\usepackage{graphicx} 
\usepackage{natbib}  
\usepackage{caption} 
\DeclareCaptionStyle{ruled}{labelfont=normalfont,labelsep=colon,strut=off} 
\usepackage{algorithm}
\usepackage{algorithmic}

\usepackage{newfloat}
\usepackage{listings}
\floatstyle{ruled}
\newfloat{listing}{tb}{lst}{}
\floatname{listing}{Listing}
\usepackage{subcaption}
\usepackage{xcolor}
\usepackage[export]{adjustbox}
\usepackage{amssymb}
\usepackage{amsmath}

\usepackage{amsmath,amsfonts,bm}

\def\eqref#1{equation~\ref{#1}}

\def\1{\bm{1}}

\def\rz{{\textnormal{z}}}

\DeclareMathAlphabet{\mathsfit}{\encodingdefault}{\sfdefault}{m}{sl}
\SetMathAlphabet{\mathsfit}{bold}{\encodingdefault}{\sfdefault}{bx}{n}

\newcommand{\E}{\mathbb{E}}

\newcommand{\KL}{D_{\mathrm{KL}}}

\usepackage{nidanfloat}

\usepackage{pifont}
\newcommand{\cmark}{\ding{51}}%
\newcommand{\xmark}{\ding{55}}%

\nocopyright 

\title{Curiosity-Driven Exploration via Latent Bayesian Surprise
}
\author{
    Pietro Mazzaglia,
    Ozan Catal,
    Tim Verbelen,
    Bart Dhoedt
}
\affiliations{
    IDLab, Ghent University

    \{firstname\}.\{lastname\}@ugent.be
}

\begin{document}

\maketitle

\begin{abstract}
The human intrinsic desire to pursue knowledge, also known as curiosity, is considered essential in the process of skill acquisition. With the aid of artificial curiosity, we could equip current techniques for control, such as Reinforcement Learning, with more natural exploration capabilities. A promising approach in this respect has consisted of using Bayesian surprise on model parameters, i.e. a metric for the difference between prior and posterior beliefs, to favour exploration. In this contribution, we propose to apply Bayesian surprise in a latent space representing the agent’s current understanding of the dynamics of the system, drastically reducing the computational costs. We extensively evaluate our method by measuring the agent's performance in terms of environment exploration, for continuous tasks, and looking at the game scores achieved, for video games. Our model is computationally cheap and compares positively with current state-of-the-art methods on several problems. We also investigate the effects caused by stochasticity in the environment, which is often a failure case for curiosity-driven agents. In this regime, the results suggest that our approach is resilient to stochastic transitions.
\end{abstract}

\noindent \section{Introduction}
\label{sec:intro}

Agents can be trained with Reinforcement Learning (RL) to successfully accomplish tasks by maximising a reward signal that encourages correct behaviors and penalizes wrong actions. For instance, agents can learn to play video games by maximizing the game score \citep{Mnih2015DQNature} or achieve robotic manipulation tasks, such as solving a Rubik's cube \citep{OpenAI2019SolvingRubik}, by following human-engineered rewards. However, how to correctly define reward functions to develop general skills remains an unsolved problem, and it is likely to stumble across undesired behaviours when designing rewards for complex tasks  \citep{Amodei2016RewardHacking, Clark2016FaultyReward, Krakovna2020SpecificationGaming, Popov2017LegoStacking}.

In contrast to RL agents, humans can learn behaviors without any external rewards, due to the intrinsic motivation that naturally drives them to be active and explore the environment \citep{Larson2011IM, Legault2016IntrExtrMotivation}. 
The design of similar mechanisms for RL agents opens up possibilities for training and evaluating agents without external rewards \citep{Matusch2020EvalNoRew}, fostering more self-supervised strategies of learning.

The idea of instilling intrinsic motivation, or `curiosity', into artificial agents has raised a large interest in the RL community \citep{Oudeyer2007IntrinsicMotivationSystems, Schmidhuber1991CuriousModel}, where curiosity is used to generate intrinsic rewards that replace or complement the external reward function. 
However, what is the best approach to generate intrinsic bonuses is still unsettled and current techniques underperform in certain domains, such as stochastic or ambiguous environments \citep{Wauthier2021Mouse}.

Several successful approaches modeled intrinsic rewards as the `surprisal' of a model. In layman's terms, this can be described as the difference between the agent's belief about the environment state and the ground truth, and can be implemented as the model's prediction error \citep{Achiam2017SurpriseDeepRL, Pathak2017ICM}. However, searching for less predictable states suffers from the `NoisyTV problem', where watching a screen outputting white random noise appears more interesting than other exploratory behaviours \citep{Schmidhuber2010FormalTheoryCreativity}. This because the noise of the TV is stochastic and thus results generally more interesting than the rest of the environment \citep{Burda2019LSCuriosity}.

In contrast, Bayesian surprise \citep{Laurent2006BayesianSurprise} measures the difference between the posterior and prior beliefs of an agent, after observing new data. As we also show in this work, this means that for stochastic transitions of the environment, which carry no novel information to update the agent's beliefs, low intrinsic bonuses are provided, potentially overcoming the NoisyTV issue. Previous work adopting Bayesian surprise for exploration has mostly focused on evaluating surprise in the model's parameter space \citep{Houthooft2016VIME}, which suffers from being computationally-expensive. 

\textbf{Contributions.} In this work, we present a new curiosity bonus based on the concept of Bayesian surprise. Establishing a latent variable model in the task dynamics, we derive \textbf{L}atent \textbf{B}ayesian \textbf{S}urprise (\textbf{LBS}) as the difference between the posterior and prior beliefs of a latent dynamics model. Our dynamics model 
uses the random variable in latent space to predict the future, while at the same time capturing any uncertainty in the dynamics of the task. 


The main contributions of the work are as follows: (\textit{i}) a latent dynamics model, which captures the information about the dynamics of the environment in an unobserved variable that is used to predict the future state, (\textit{ii}) a new Bayesian surprise inspired exploration bonus, derived as the information gained with respect to the latent variable in the dynamics model, (\textit{iii}) evaluation of the exploration capabilities on several continuous-actions robotic simulation tasks and on discrete-actions video games, and comparison with other exploration strategies, and (\textit{iv}) assessment of the robustness to stochasticity, by comparing to the other baselines on tasks with stochastic transitions in the dynamics.

The results empirically show that our LBS method either performs on par and often outperforms state-of-the-art methods, when the environment is mostly deterministic, making it a strongly valuable method for exploration. Furthermore, similarly to methods using Bayesian surprise in parameter space, LBS is resilient to stochasticity, and actually explores more in-depth than its parameter space counterparts in problems with stochastic dynamics, while also being computationally cheaper. Further visualization, is available on the project webpage.\footnote{\url{https://lbsexploration.github.io/}}

\section{Background}
\label{sec:background}

We focus on exploration bonuses to incentivize exploration in RL. To foster the reader's understanding, we first introduce standard notation and common practices. 

\textbf{Markov Decision Processes.} The RL setting can be formalized as a Markov Decision Process (MDP), which is denoted with the tuple $\mathcal{M} = \{\mathcal{S}, \mathcal{A}, T, R, \gamma\}$, where $\mathcal{S}$ is the set of states, $\mathcal{A}$ is the set of actions, $T$ is the state transition function, also referred to as the dynamics of the environment, $R$ is the reward function, which maps transitions into rewards, and $\gamma$ is a discount factor. The dynamics of the task can be described as $p(s_{t+1}|s_t,a_t)$ that is the probability that action $a_t$ brings the system to state $s_{t+1}$ from state $s_t$, at the next time step $t+1$. The objective of the RL agent is to maximize the expected discounted sum of rewards over time, also called return, and indicated as ${G_t = \sum_{k=t+1}^T \gamma^{(k-t-1)}r_k}$.



\textbf{Policy Optimization.} In order to maximize the returns, the agent should condition its actions on the environment's current state. The policy function $\pi(a_t|s_t)$ is used to represent the probability of taking action $a_t$ when being in state $s_t$. Several policy-optimization algorithms also evaluate two value functions, $V(s_t)$ and $Q(s_t,a_t)$, to estimate and predict future returns with respect to a certain state or state-action pair, respectively.    


\textbf{Intrinsic Motivation.} Curious agents are designed to search for novelty in the environment and to discover new behaviours, driven by an intrinsically motivated signal. Practically, this comes in the form of self-generated rewards $r^{(i)}$ that can complement or replace the external rewards $r^{(e)}$ of the environment. The combined reward at time step $t$ can be represented as: $r_t = \eta_e r^{(e)}_t + \eta_i r^{(i)}_t$, where $\eta_e$ and $\eta_i$ are factors adopted to balance external and intrinsic rewards. How to optimally balance between exploration with intrinsic motivation and exploitation of external rewards is still an unanswered question, which we do not aim to address with our method. Instead, similarly to what done in other works \citep{Shyam2019MAX, Burda2019LSCuriosity, Pathak2019Disagreement, Ratzlaff2020ImplGenModelExpl, Tao2020NSRS}, we focus on the exploration behaviour emerging from the self-supervised intrinsic motivation signal. 

\textbf{Surprisal and Bayesian Surprise.} The surprisal, or information content, of a random variable is defined as the negative logarithm of its probability distribution. In an MDP, at time step $t$, we can define the surprisal with respect to next-step state as $-\log p(s_{t+1}|s_t,a_t)$. By using a model with parameters $\theta$ to fit the transition dynamics of the task, we can define surprisal in terms of the probability estimated by the model, namely $- \log p_\theta(s_{t+1}|s_t, a_t)$. Such surprisal signal has been adopted for exploration in several works \citep{Achiam2017SurpriseDeepRL,Pathak2017ICM}. One shortcoming of these methods is that a stochastic transition, e.g. rolling a die, will always incur into high surprisal values, despite the model having observed the same transition several times. This problem has been treated in literature as the `NoisyTV problem' \citep{Schmidhuber2009NoisyTV, Schmidhuber2010FormalTheoryCreativity}.

In contrast, Bayesian surprise \citep{Laurent2006BayesianSurprise} can be defined as the information gained about a random variable, by observing another random variable. For instance, we can compute the information gained about the parameters of the model $\theta$ by observing new states as $\mathcal{I}(\theta; s_{t+1}|s_t, a_t)$. Such signal has been used for exploration exploiting Bayesian neural networks \citep{Houthooft2016VIME}, where Bayesian surprise is obtained by comparing the weights distribution before and after updating the model with newly collected states. However, this procedure is extremely expensive, as it requires an update of the model for every new transition. Alternatively, an approximation of Bayesian surprise is obtainable by using the variance of an ensemble of predictors \citep{Pathak2019Disagreement,Sekar2020Plan2Explore}, though this method still requires to train several models.



\section{Latent Bayesian Surprise}
\label{sec:method}

Our method provides intrinsic motivation through a Bayesian surprise signal that is computed with respect to a latent variable. First, we describe how the latent dynamics model works and how it allows the computation of Bayesian surprise in latent space. Then, we present an overview of the different components of our model and explain how they are concurrently trained to fit the latent dynamics, by exploiting variational inference. Finally, we show how the intrinsic reward signal for LBS is obtained from the model's predictions and discuss connections with other methods.

\begin{figure}[htb]
    \centering
    \begin{subfigure}[b]{0.49\columnwidth}
         \centering
         \includegraphics[width=0.97\linewidth]{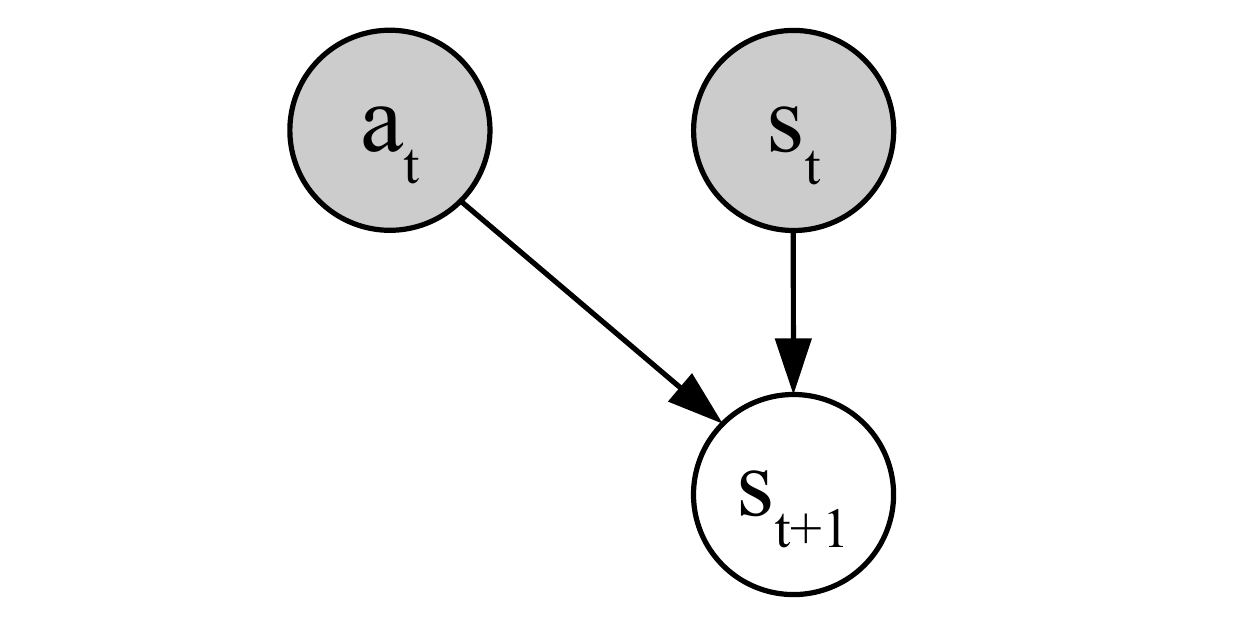}
         \caption{MDP dynamics}
         \label{subfig:mdp}
    \end{subfigure}
    \begin{subfigure}[b]{0.49\columnwidth}
         \centering
         \includegraphics[width=0.97\linewidth]{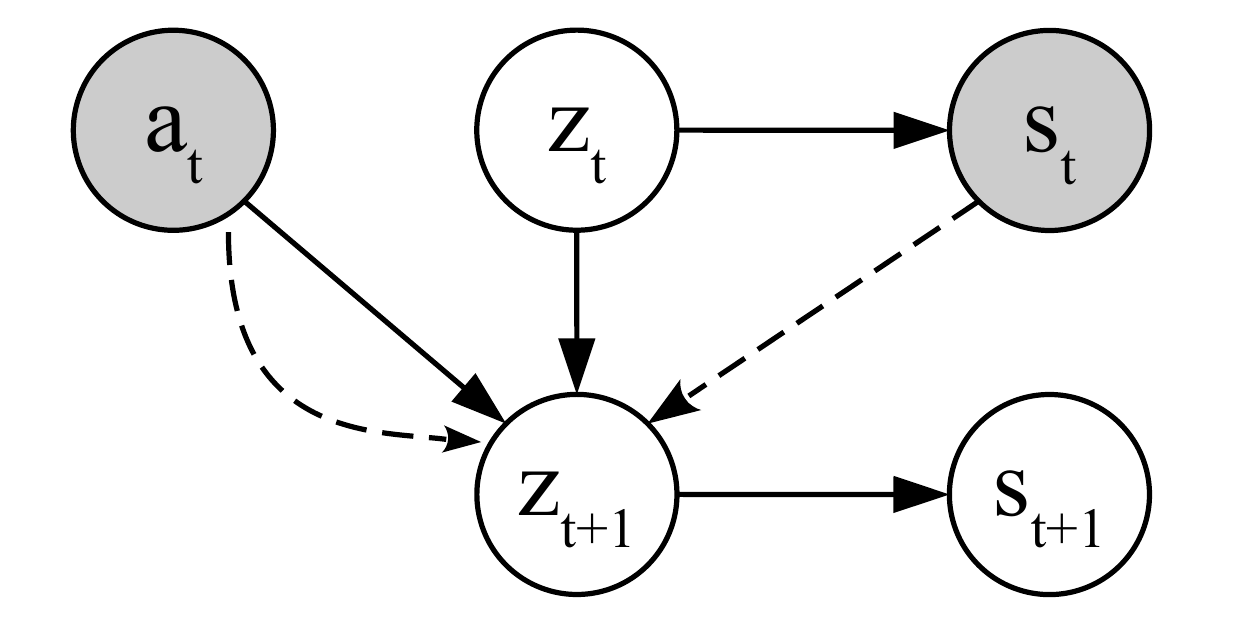}
         \caption{Latent dynamics (ours)}
     \label{subfig:latent_mdp}
    \end{subfigure}
    \caption{\textbf{Dynamics graphical models.} The model observes $s_t$ and $a_t$. Solid lines indicate generative processes and dashed lines indicate the inference ones.}
    \label{fig:mdp_models}
\end{figure}

\textbf{Latent Dynamics.} The transition dynamics of an MDP can be summarized as the probability of the next state, given the current state and the action taken at the current time step, namely $p(s_{t+1}|s_t, a_t)$. The associated generative process is presented in Figure \ref{subfig:mdp}. In the case of deterministic dynamics, the next state is just a function of the current state and action. For non-deterministic dynamics, there would be a distribution over the next state, from which samples are drawn when the state-action pair is triggered. The entropy of such distribution determines the uncertainty in the dynamics.


With the aim of capturing the environment's uncertainty and to compute the Bayesian surprise given by observing new states, we designed the latent dynamics model in Figure \ref{subfig:latent_mdp}. The intermediate latent variable $z_{t+1}$ should contain all the necessary information to generate $s_{t+1}$, so that by inferring the latent probability distribution as $p(z_{t+1}|s_t, a_t)$ from previous state and action, we can then estimate future state probability as $p(s_{t+1}|z_{t+1})$. 

As we discuss later in this Section, we can train a model to maximize an evidence lower bound on the future states likelihood that matches our latent variable model. Then, the most appealing aspect for exploration is that we can now compute Bayesian surprise in latent space as $\mathcal{I}(z_{t+1}; s_{t+1}|s_t, a_t)$, which is the information gained with respect to the latent variable by observing the actual state.  

\begin{figure}[b]
     \centering
     \includegraphics[width=\linewidth]{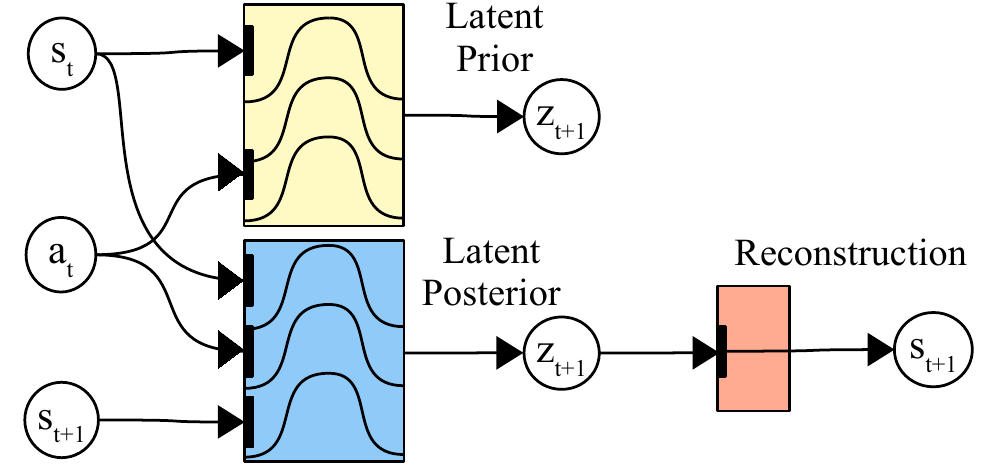}
     \caption{\textbf{LBS overview.} The modules of LBS, with input and output variables. Latent Prior and Posterior output distributions, while the Reconstruction model outputs point estimates.}
     \label{fig:dynamics_model}
\end{figure}

\textbf{Model Overview.} A dynamics model with parameters $\theta$ can be trained to match the environment dynamics (as in Figure \ref{subfig:mdp}) by maximizing the log-likelihood $\log p(s_{t+1}|s_t,a_t)$ of its predictions.

Similarly, given our latent variable model, we can train a dynamics model to maximize an evidence lower bound on the log-likelihood of future states. For this purpose, the LBS model is made of the following components:
\begin{equation*}
    \begin{split}
        & \textrm{Latent Prior:} \\
        & \textrm{Latent Posterior:} \\
        & \textrm{Reconstruction model:} \\
    \end{split}
    \qquad \quad
    \begin{split}
        &  p_\theta(z_{t+1}|s_t, a_t), \\
        &  q_\theta(z_{t+1}|s_t, a_t, s_{t+1}), \\
        &  p_\theta(s_{t+1}|z_{t+1}), \\
    \end{split}
\end{equation*}

\noindent which are displayed in Figure \ref{fig:dynamics_model}. The latent prior component represents prior beliefs over the next state's latent variable. The latent posterior $q(z_{t+1})$ represents a variational distribution that approximates the true posterior of the latent variable, given the observed data $s_{t+1}$. Finally, the reconstruction module allows to generate the next state from the corresponding latent. Overall, the model resembles a conditional VAE \cite{Kingma2014VAE}, trained to autoencode the next states, conditioned on current states and actions.

All the components parameters $\theta$ are jointly optimized by maximizing the following variational lower bound on future states log-likelihood: 
\begin{equation}
\begin{split}
    \mathcal{J} = &\E_{\rz_{t+1}\sim q(z)} [ \log p_\theta(s_{t+1}|z_{t+1})] \\ 
    &- \beta \KL [q_\theta(z_{t+1}|s_t, a_t,s_{t+1}) \Vert p_\theta(z_{t+1}|s_t, a_t) ]
    \label{eq:elbo}
\end{split}
\end{equation}
where $\beta$ 
is introduced 
to control disentanglement in the latent representation, as in \citep{Higgins2017BetaVAE}. The derivation of the objective is available in the Appendix. 

\textbf{Intrinsic Rewards.} 
In our method, we are interested in measuring the amount of information that is gained by the model when facing a new environment's transition and using that as an intrinsic reward to foster exploration in RL. Every time the agent takes action $a_t$ while being in state $s_t$, it observes a new state $s_{t+1}$ that completes the transition and brings new information to the dynamics model. 

Such information gain can be formulated as the KL divergence between the latent prior and its approximate posterior and adopted as an intrinsic reward for RL as follows:
\begin{equation}
\begin{split}
    r^{(i)}_t &= \mathcal{I}(z_{t+1};s_{t+1}|s_t, a_t) \\
    &\approx \KL [q_\theta(z_{t+1}|s_t, a_t, s_{t+1}) \Vert p_\theta(z_{t+1}|s_t, a_t) ]
\end{split}
\end{equation}
The above term can be efficiently computed by comparing the distributions predicted by the latent prior and the latent posterior components. The signal provided should encourage the agent to collect transitions where the predictions are more uncertain or erroneous. 

The intrinsic motivation signal of LBS can also be reformulated as (conditioning left out for abbreviation):
\begin{equation}
\begin{split}
    & \KL [q_\theta(z_{t+1}) \Vert p_\theta(z_{t+1}) ] = \\
    &= \E_{q_\theta(z_{t+1})}[ \log q_\theta(z_{t+1}) - \log p_\theta(z_{t+1}) ]  \\
    &= -H[q_\theta(z_{t+1})] + H[q_\theta(z_{t+1}),p_\theta(z_{t+1})]
\end{split}
\end{equation}
where the left term is the entropy of the latent posterior and the right term is the cross-entropy of $p$ relatively to $q$. Maximizing our bonus can thus be interpreted as searching for states with minimal entropy of the posterior and a high cross-entropy value between the posterior and the prior. 

Assuming the LBS posterior correctly approximates the true posterior of the system dynamics, the cross-entropy term closely resembles the `surprisal' bonus adopted in other works \citep{Achiam2017SurpriseDeepRL, Pathak2017ICM, Burda2019LSCuriosity}.  Using LBS can then be seen as maximizing the `surprisal', while trying to avoid high-entropy, stochastic states.

\begin{figure*}[htb!]
    \centering
    \includegraphics[width=\textwidth]{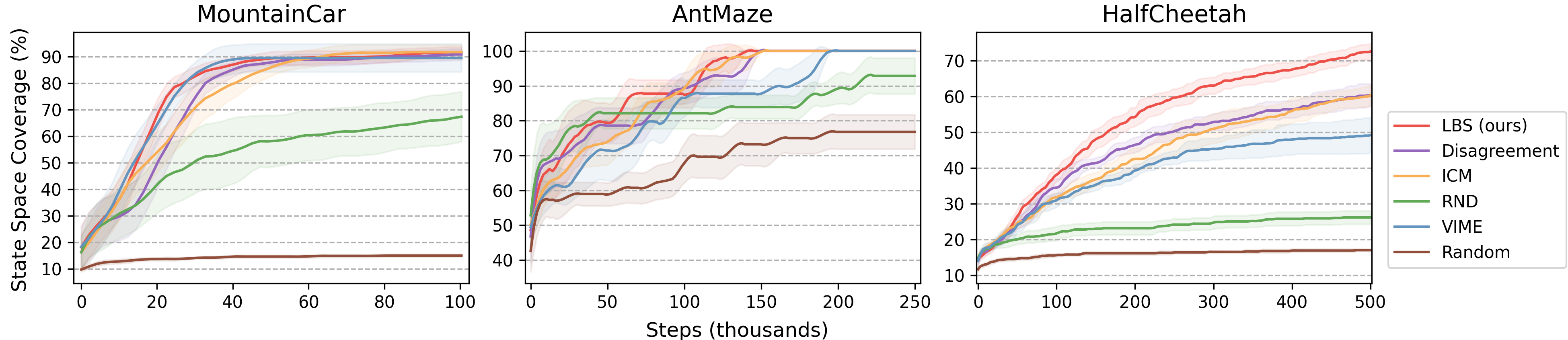}
    \caption{\textbf{Continuous Control results.} A comparison of our method against several baselines on continuous control tasks. Lines show the average state-space coverage (standard deviations in shade) in terms of percentage of bins visited by the agents.} 
    \label{fig:continuous_results}
\end{figure*}

\section{Experiments}
\label{sec:results}

The aim of the experiments is to compare the performance of the LBS model and intrinsic rewards against other approaches for exploration in RL.  

\textbf{Environments.} Main results are presented with respect to three sets of environments: continuous control tasks, discrete-action games, and tasks with stochastic transitions. The continuous control tasks include the classic Mountain Car environment \citep{Moore1990MountainCar}, the Mujoco-based Half-Cheetah environment \citep{Todorov2012Mujoco}, and the Ant Maze environment used in \citep{Shyam2019MAX}. The discrete-action games include 8 video games from the Atari Learning Environment (ALE; \citet{Bellemare2013ALE}) and the Super Mario Bros. game, which is a popular NES platform game. The stochastic tasks include an image-prediction task with stochastic dynamics and two stochastic variants of Mountain Car, including a NoisyTV-like component. 

In this Section, we consider curious agents that only optimize their self-supervised signal for exploration. This means that we omit any external rewards, by setting $\eta_e = 0$ (see Background). 
This focuses the agents solely on the exploratory behaviors inspired by the curiosity mechanisms. For all tasks, we update the policy using the Proximal Policy Optimization algorithm (PPO; \citet{Schulman2017PPO}). For all model's components, we use neural networks. For the model's latent stochastic variable, we use distributional layers implemented as linear layers that output the means and standard deviations of a multivariate gaussian. 

\textbf{Zero-shot Adaptation.} We present additional experiments on the Deep Mind Control suite \cite{Tassa2018DMC} in the Appendix. As in Plan2Explore \cite{Sekar2020Plan2Explore}, we use intrinsic motivation to train an exploration policy, which collects data to improve the agent's model. Then, the model is used to train an exploitative policy on the environment's rewards and its zero-shot performance is evaluated. In these visual control tasks, we show that the intrinsic motivation bonus of LBS combines well with model-based RL, achieving similar or higher performance than Plan2Explore and requiring no additional predictors to be trained. 


\subsection{Continuous Control}


In our continuous control experiments, we discretize the state-space into bins and compare the number of bins explored, in terms of coverage percentage. An agent being able to visit a certain bin corresponds to the agent being able to solve an actual task that requires reaching that certain area of the state space. Thus, it is important that a good exploration method would be able to reach as many bins as possible.  

We compare against the following baselines:
\begin{itemize}
\item \textit{Disagreement} \citep{Pathak2019Disagreement}: an ensemble of models is trained to match the environment dynamics. The variance of the ensemble predictions is used as the curiosity signal.
\item \textit{Intrinsic Curiosity Model} (ICM; \citet{Pathak2017ICM}): intrinsic rewards are computed as the mean-squared error (MSE) between a dynamics model's predictions in feature space and the true features. States are processed into features using a feature network, trained jointly with the model to optimize an inverse-dynamics objective.
\item \textit{Random Network Distillation} (RND; \citet{Burda2019RND}): features are obtained with a fixed randomly initialized neural network. Intrinsic rewards for each transition are the prediction errors between next-state features and the output of a distillation network, trained to match the outputs of the random feature network.
\item \textit{Variational Information Maximizing Exploration} (VIME; \citet{Houthooft2016VIME}): the dynamics is modeled as a Bayesian neural network (BNN; \citet{Bishop1997BNN}). Intrinsic rewards for single transitions are shaped as the information gain computed with respect to the BNN's parameters before and after updating the network, using the new transition's data.
\item \textit{Random}: an agent that explores by performing a series of random actions. Note that employing random actions is equivalent to having a policy with maximum entropy of actions, for each state. Thus, despite its simplicity, the random baseline provides a metric of how in-depth do maximum entropy RL methods explore, when receiving no external rewards \citep{Haarnoja2018SAC}.   
\end{itemize}


We found LBS to be working best in this benchmark, as it explores the most in-depth and the most efficiently in all tasks. The training curves are presented in Figure \ref{fig:continuous_results}, averaging over runs with eight different random seeds. Further comparison against RIDE \cite{Raileanu2020RIDE} and NGU \cite{Badia2020NGU}, which employ episodic counts to modulate exploration, are presented in Appendix.
 
\textbf{Mountain Car.} In Mountain Car, the two-dimensional state space is discretized into 100 bins. Figure \ref{fig:continuous_results} shows that LBS, VIME, ICM and Disagreement all reach similar final performance, with around 90\% of coverage. In particular, LBS and VIME are on average faster at exploring in the first 30k steps. RND struggles behind with about 67\% of visited bins, doing better only than the Random baseline ($\sim$15\%).

\textbf{Ant Maze.} In the Ant Maze environment, the agent can explore up to seven bins, corresponding to different aisles of a maze. LBS, ICM and Disagreement perform best in this environment, reaching the end of the maze in all runs and before 150k steps. VIME also reaches 100\% in all runs but takes longer. RND and the Random baselines saturate far below 100\% coverage.

\textbf{Half-Cheetah.} In the Half-Cheetah environment, the state space is discretized into 100 bins. In this task, which has the most complex dynamics compared to the others, LBS reaches the highest number of bins, with  around 73\% of coverage. ICM and Disagreement follow with $\sim$60\%, and VIME with $\sim$49\%. RND lacks behind by doing slightly better than the Random baseline ($\sim$26\% vs $\sim$17\%). 


\begin{figure*}
    \centering
    \includegraphics[width=0.9\textwidth]{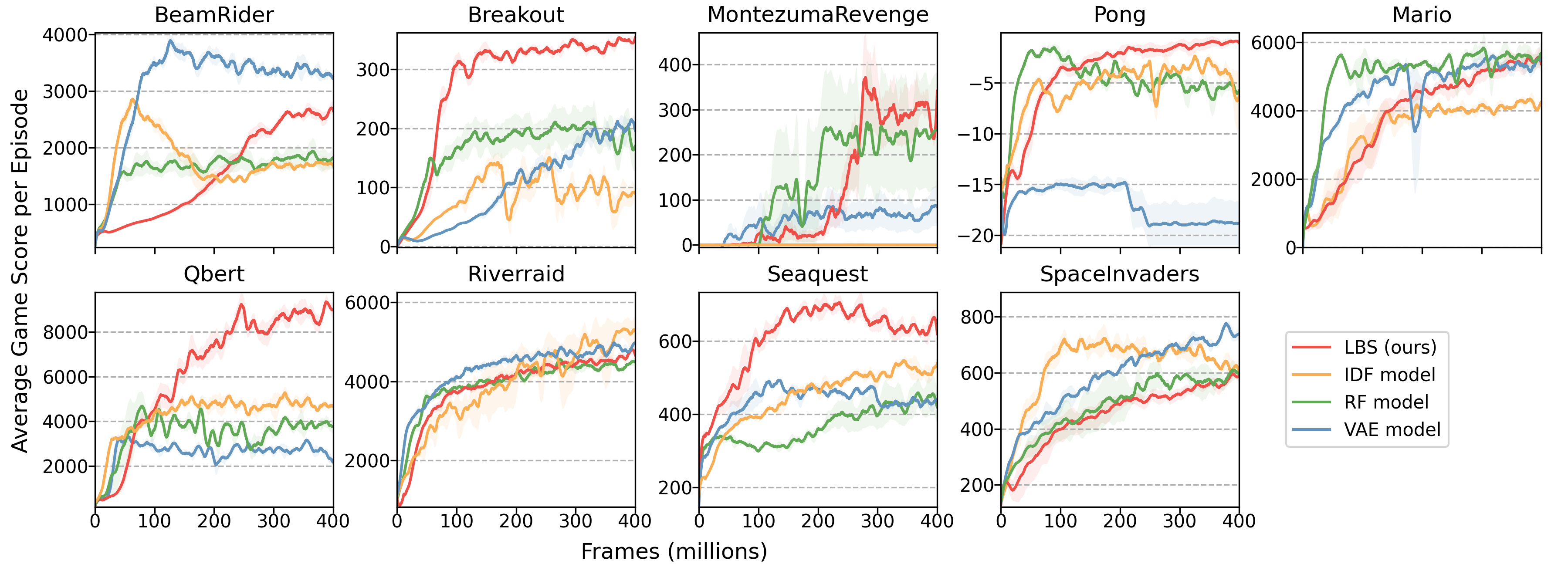}
    \caption{\textbf{Arcade Games results.} A comparison of LBS against surprisal-based models, using different sets of features, on 8 selected Atari and the Super Mario Bros. games. Lines show the average game score per episode (standard deviations in shade).} 
    \label{fig:arcade_results}
\end{figure*}

\subsection{Arcade Games} 

For the arcade games, the environments chosen are designed in a way that either requires the player to explore in order to succeed, e.g. Qbert, or to survive as long as possible to avoid boredom, e.g. Pong. For this reason, agents are trained only with curiosity but evaluated on the game score they achieve in one episode, or, in the case of Super Mario Bros., on the distance traveled from the start position. Higher scores in this benchmark translate either into an higher number of enemies killed, objects collected, or transitions/areas/levels of the game explored, meaning that methods that perform better on this benchmark are more likely to discover meaningful skills in the environment. Combining curiosity with environment's rewards, performance in these games could be significantly improved with respect to using only curiosity but we do not compare to that setting in order to completely focus on the exploration performance. 

We follow the setup of \citep{Burda2019LSCuriosity} and compare against their baselines, which use MSE prediction error in feature space as the intrinsic motivation signal, aka surprisal in feature space. The feature space is obtained by projecting states from the environment into a lower-dimensional space using a feature model, i.e. next-state features can be expressed as $\phi_{t+1} = f(s_{t+1})$.

The different baselines use different feature models, so that the Variational Autoencoder, or \textit{VAE model}, trains an autoencoder, as in \citep{Kingma2014VAE}, concurrently with the dynamics model; the Random Features, or \textit{RF model}, uses a randomly initialized network; the  Inverse Dynamics Features, or \textit{IDF model}, uses features that allow to model the inverse dynamics of the environment.

For LBS, we also found that working in a reduced feature space, compared to the high-dimensional pixel space, is beneficial. For this purpose, we project the states from the environment into a low-dimensional feature space using a randomly initialized network, similarly to the RF model. We believe more adequate features than random could be found, though we leave this idea for future studies. 
In this setup, the  reconstruction model predicts next-state features instead of next-state pixels:
\begin{equation*}
    \begin{split}
        & \textrm{Reconstruction model:} \\
    \end{split}
    \qquad \quad
    \begin{split}
        &  p_\theta(\phi_{t+1}|z_{t+1}). \\
    \end{split}
\end{equation*}
A performance comparison between using pixel and feature reconstruction is provided in the Appendix.


The training curves are shown in Figure \ref{fig:arcade_results}, presenting the original results from \citep{Burda2019LSCuriosity} for the baselines and an average of five random seed runs for LBS.
The empirical results are favorable towards the LBS model, which achieves the best average final score in 5 out 9 games: Montezuma Revenge, Pong, Seaquest, Breakout, and Qbert, with a large margin for the latter three; and performs comparably to the other baselines in all other games.




\subsection{Stochastic Environments} 

Our stochastic benchmark is composed of three tasks: an image-prediction task, where we quantitatively assess the intrinsic rewards assigned by each method for deterministic and stochastic transitions, and two stochastic variants of the Mountain Car control problem, presenting an additional state that is randomly controlled by an additional action. The additional low-dimensional state in Mountain Car can be seen as a one-pixel NoisyTV that is controlled by the additional action's remote.  

\begin{figure}[t]
    \centering
    \begin{subfigure}[b, valign=c]{0.4\columnwidth}
         \centering
         \includegraphics[width=0.97\linewidth]{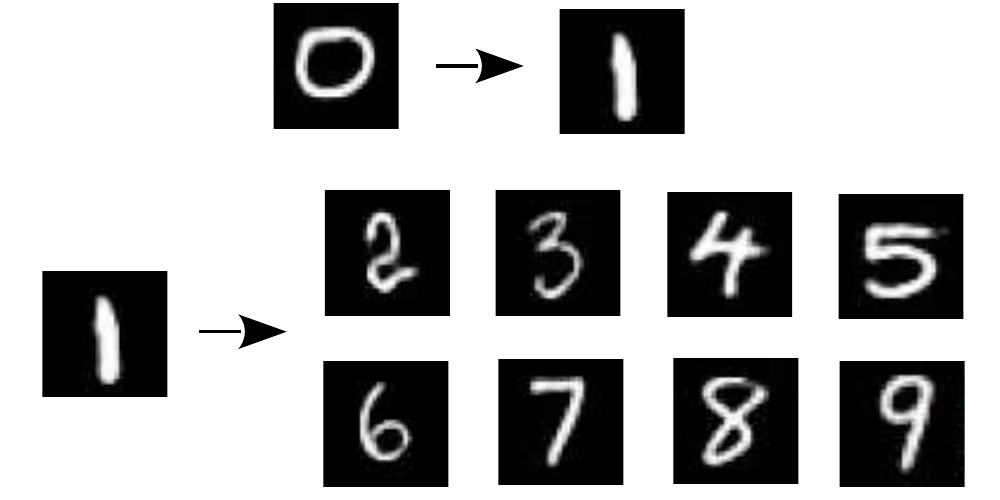}
         \caption{Task Dynamics}
         \label{subfig:mnist_task}
    \end{subfigure}
    \begin{subfigure}[b, valign=c]{0.58\columnwidth}
         \centering
         \includegraphics[width=0.97\linewidth]{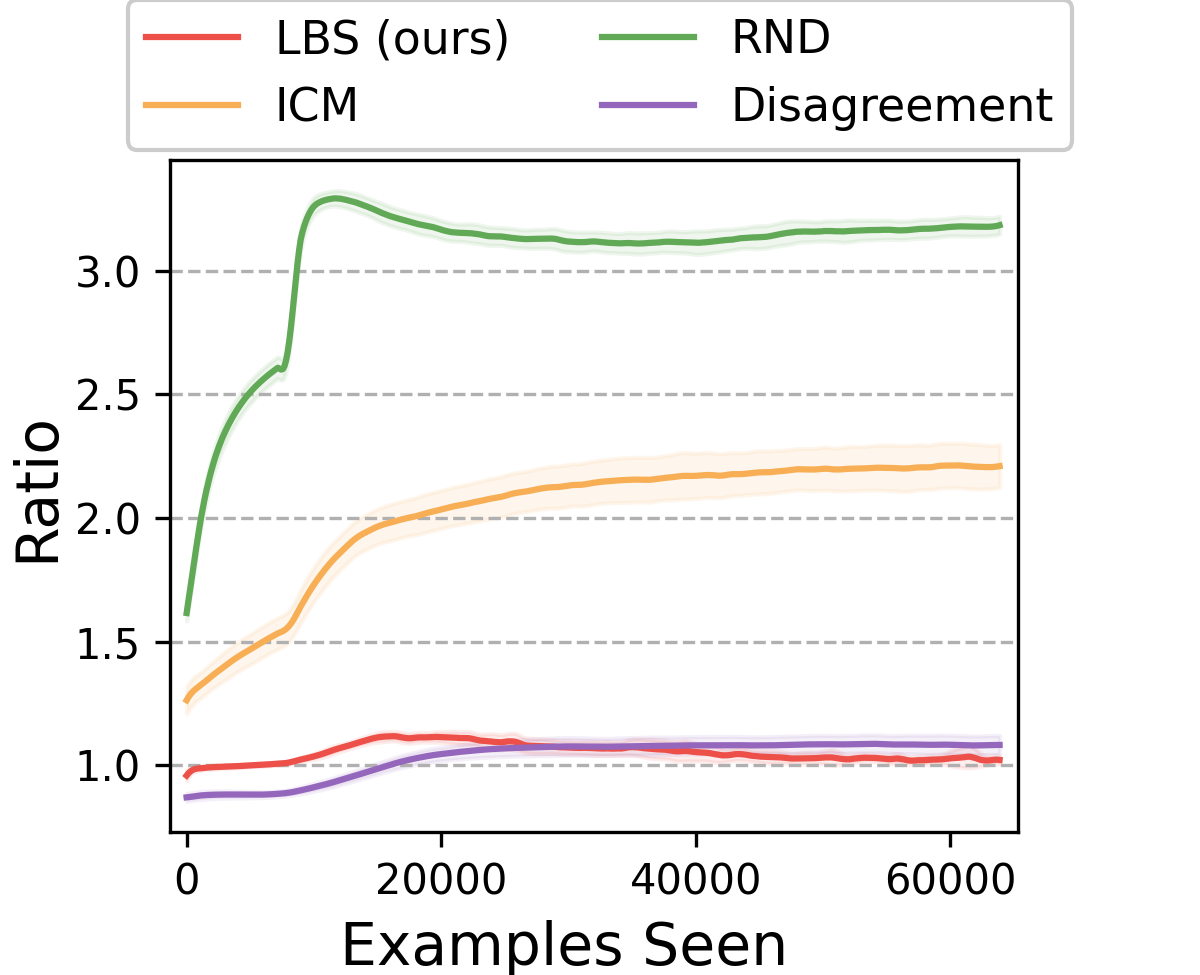}
         \caption{Baselines comparison}
     \label{subfig:mnist_baselines}
    \end{subfigure}
    \caption{\textbf{Stochastic MNIST.} (a) Image prediction stochastic task based on the MNIST dataset samples. (b) Average intrinsic motivation ratio over training samples, in ten runs. The closer the ratio to the unity, at convergence, the better.}
    \label{fig:stochastic}
\end{figure}

\textbf{Image Task.} Similarly to \citep{Pathak2019Disagreement}, we employ the Noisy MNIST dataset \citep{lecun1995MNIST} to perform an experiment on stochastic transitions. Taking examples from the test set of MNIST, we establish a fictitious dynamics that always starts either from an image of a zero or a one: a 0-image always transitions to a 1-image, while a 1-image transitions into an image representing a digit between two and nine (see Figure \ref{subfig:mnist_task}). 

We assess the performance in terms of the ratio between the intrinsic motivation provided for transitions starting from 1-images and transitions starting from 0-images. After having seen several samples starting from the 1-image, the agent should eventually understand that results associated with this more stochastic transition do not bring novel information about the task dynamics, and should lose interest with respect to it. Thus, the expected behavior is that the ratio should eventually lean to values close to the unity.

We train the models uniformly sampling random transitions in batches of 128 samples and run the experiments with ten random seeds. In Figure \ref{subfig:mnist_baselines}, we compare LBS to Disagreement, ICM and RND. We observe that LBS and Disagreement are the only methods that eventually overcome the stochasticity in the transitions starting from 1-images, maintaining a ratio of values close to one at convergence. Both ICM and RND, instead, keep finding the stochastic transition more interesting at convergence. 


\begin{figure*}[b!]
     \centering
     \includegraphics[width=0.92\columnwidth, valign=c]{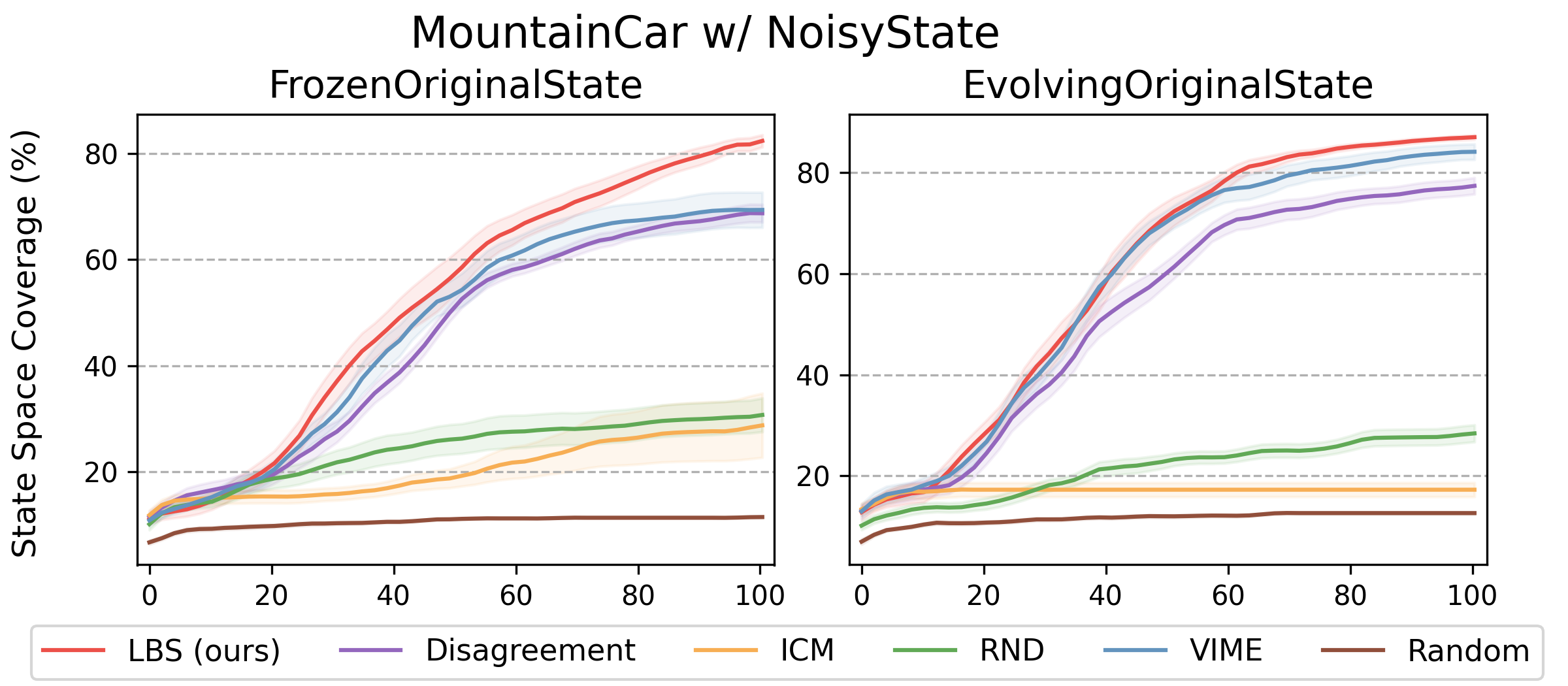}
    \resizebox{1.05\columnwidth}{!}{
    \begin{tabular}[valign=c]{c|ccc|cc}
    \multicolumn{1}{c}{}  &\multicolumn{3}{|c}{State Space Coverage (\%)} &\multicolumn{2}{|c}{Reduction (\%)} \\
    \multicolumn{1}{c}{}  & \multicolumn{1}{|c}{NoStoch} &\multicolumn{1}{c}{Frozen} &\multicolumn{1}{c}{Evolving} &\multicolumn{1}{|c}{Frozen} &\multicolumn{1}{c}{Evolving}
    \\[0.25em] \hline
    LBS (ours)         & \textbf{91.75} & \textbf{82.38} & \textbf{87.0} & \textbf{{\color{red} -10.21 }}
  & \textbf{{\color{red} -5.18 }} \\
    Disagreement             & 90.88 & 68.75 & 77.38 & {\color{red} -24.35} & {\color{red} -14.85} \\
    ICM             & \textbf{91.75} & 28.75 & 17.25 & {\color{red} -68.66} & {\color{red} -81.20} \\
    RND             & 67.38 & 30.75  & 28.38 & {\color{red} -54.36} & {\color{red} -57.88}   \\
    VIME             & 89.57 & 69.38  & 84.12 &  {\color{red} -22.54} & {\color{red} -6.08} \\
    Random             & 15.0 & 11.5  & 12.62 & {\color{red} -23.33} & {\color{red} -15.87}  \\
    \end{tabular}}
     \caption{\textbf{Stochastic Mountain Car.} On the left, training curves on the two variants of the stochastic Mountain Car problem are displayed, showing the average state space coverage over eight random seeds (standard deviations in shade). On the right, the table compares the final performance with the original non-stochastic environment, highlighting the reductions in performance.}
     \label{fig:stoch_mountain_car}
\end{figure*}

\textbf{Stochastic Mountain Car.} The original Mountain Car continuous control problem is made of a two-dimensional state space, position and velocity of the car, which we refer to as the OriginalState, and a one-dimensional action space, controlling the force to apply to the car to move. We extended the environment to be stochastic by adding a one-dimensional state, referred to as the NoisyState, and a one-dimensional action, ranging from $[-1,1]$ which works as a remote for the NoisyState. When this action's value is higher than 0, the remote is triggered, updating the NoisyState value, by sampling uniformly from the $[-1, 1]$ interval. Otherwise, the task works like the standard Mountain Car, and the agent can explore up to 100 bins.  

We experiment with two versions of the environment:
\begin{itemize}
    \item \textit{FrozenOriginalState}: when the remote is triggered, the OriginalState is kept frozen, regardless of the force applied to the car. This allows the agent to focus on the NoisyState changes, whilst not losing the velocity and the momentum of the car. 
    \item \textit{EvolvingOriginalState}: when the remote is triggered, the OriginalState is updated but the force applied is zero. This means the agent has to decide whether giving up on the original task to focus on the NoisyState varying.
\end{itemize}
We hypothesized that, in the Frozen scenario, a surprisal-based method, like ICM, would sample the NoisyState but also widely explore the OriginalState, as the gravity normally pushing down the car is frozen when the agent is distracted by the noisy action, representing no impediment to exploration. In practice, we see that ICM's average performance on the Frozen problem is better than on the Evolving setup but is still strongly limited by stochasticity. 

Average state space coverage for several baselines is displayed in Figure \ref{fig:stoch_mountain_car}. As also highlighted in the Figure's table, LBS remains the best performing method in both the variants of the stochastic environment, being strongly robust to NoisyTV-like stochasticity. Disagreement and VIME also show to be resilient to stochasticity, though exploring less than LBS. Both ICM and RND's performance are strongly undermined by the randomness in the task. The tabular results also show that LBS is the method that least reduced its exploration performance, compared to the original non-stochastic Mountain Car experiment and that ICM is the method that suffered the presence of noise the most.

\begin{table*}[t!]
    \small
    \centering
    \caption{We summarize and compare several exploration methods, highlighting similarities and differences.}
    \begin{tabular}{cccccc}
    \hline
    Algorithm & Objective & Model Loss & Distributions & Ensemble & Episodic \\ \hline
    ICM      & $-\log p(\phi_{t+1}|\phi_t, a_t)$ & Forward + Inverse Dynamics & \xmark & \xmark & \xmark \\
    RND & $-\log p(\phi_{t+1}|s_{t+1})$ & Knowledge Distillation  & \xmark & \xmark & \xmark \\
    VIME      & $\KL[q(\theta'|s_t, a_t)\Vert q(\theta|s_t, a_t)]$ & ELBO (variational weights)  & \cmark \rlap{$^{\text{(weights $\theta$)}}$} & \xmark & \xmark \\
    Disagreement & $\approx IG(s_{t+1}; \theta_{1:k}|s_t, a_t)$ & Forward Dynamics (Ensemble)  & \xmark & \cmark & \xmark \\
    Plan2Explore & $\approx IG(h_{t+1}; \theta_{1:k}|s_t, a_t)$ & Forward Dynamics (Ensemble)  & \xmark & \cmark & \xmark\\
    RIDE & $ \Vert \phi_{t+1} - \phi_{t} \Vert_2 /\sqrt{N_{ep}(s_{t+1})}$ & Forward + Inverse Dynamics  & \xmark & \xmark & \cmark \\
    NGU & $\approx \alpha_t/\sqrt{N_{ep}(\phi_{t+1})}$ & Inverse Dynamics  & \xmark & \xmark & \cmark \\
    LBS (ours) & $IG(z_{t+1}; s_{t+1}|s_t, a_t)$ & ELBO (variational latent)  & \cmark\rlap{$^{\text{(latent $z$)}}$} & \xmark & \xmark \\
    \hline
    \end{tabular}\\
    $\phi=f(s)$: features; $\theta$: model parameters; $\theta'$: $\theta$ after model update; $h$: hidden state of a RNN (part of the model); $IG$: information gain; \\ $k$: ensemble models; $N_{ep}(s)$: episodic (pseudo)count of visits to $s$; $\alpha_t$: normalized RND's objective;  $z$: latent variable in the model.
    \label{tab:methods}
\end{table*}

\section{Related Work}
\label{sec:related_work}

In Table \ref{tab:methods}, we compare LBS to all the methods we benchmark against (both in main text and Appendix).

\textbf{Reinforcement Learning.} Value-based methods in RL use the Q-value function to choose the best action in discrete settings \citep{Mnih2015DQNature, Hessel2018Rainbow}. However, the Q-value approach cannot scale well to continuous environments. Policy Optimization techniques solve this by directly optimizing the policy, either learning online, using samples collected from the policy \citep{Schulman2015TRPO, Schulman2017PPO}, or offline, reusing the experience stored in a replay buffer \citep{Lillicrap2016DDPG,Haarnoja2018SAC}.

\textbf{Latent Dynamics.} In complex environments, the use of latent dynamics models has proven successful for control and long-term planning, either by using VAEs to model locally-linear latent states \citep{Watter2015E2CLatentDynamics}, or by using recurrent world models in POMDPs \citep{Buesing2018StateSpaceModels, Hafner2019Planet, Hafner2020Dreamer}.

\textbf{Intrinsic Motivation.} Several exploration strategies use a dynamics model to provide intrinsic rewards \citep{Pathak2017ICM, Burda2019RND,  Houthooft2016VIME, Pathak2019Disagreement, Kim2019EMI}. Latent variable dynamics have also been studied for exploration  \citep{Bai2020VDM, Bucher2019PerceptionBS, Tao2020NSRS}. 
Maximum entropy in the state representation has been used for exploration, through random encoders, in RE3 \cite{Seo2021RE3}, and prototypical representations, in ProtoRL \cite{Yarats2021ProtoRL}.

Alternative approaches to modelling the environment's dynamics are based on pseudo-counts \citep{Bellemare2016CountBasedExpl, Ostrovski2017CountExplNeural, Tang2017SimHash}, which use density estimations techniques to explore less seen areas of the environment, Randomized Prior Functions \citep{Osband2018RandomPriorDQN}, applying statistical bootstrapping and ensembles to the Q-value function model, or Noisy Nets \citep{Fortunato2018NoisyNets}, applying noise to the value-function network's layers. 

Some methods combine model-based intrinsic motivation with pseudo-counts, such as RIDE \cite{Raileanu2020RIDE}, which rewards the agent with for transitions that have an impact on the state representation, and NGU \cite{Badia2020NGU}, which modulates a pseudo-count bonus with the intrinsic rewards provided by RND. Remarkably, combining NGU with an adaptive exploration strategy over the agent's lifetime led Agent57 to outperform human performance in all Atari games \cite{Badia2020Agent57}.  

\textbf{Planning Exploration.} Recent breakthroughs concerning exploration in RL have also focused on using the learned environment dynamics to plan to explore. This is the case in \citep{Shyam2019MAX} and \citep{Ratzlaff2020ImplGenModelExpl}, where they use imaginary rollouts from their dynamics models to plan exploratory behaviors, and \citep{Sekar2020Plan2Explore}, where they combine a model-based planner in latent space \citep{Hafner2020Dreamer} with the Disagreement exploration strategy \citep{Pathak2019Disagreement}.

\section{Discussion}
\label{sec:discussion}

In this work, we introduced LBS, a novel approach that uses Bayesian surprise in latent space to provide intrinsic rewards for exploration in RL. Our method has proven successful in several continuous-control and discrete-action settings, providing reliable and efficient exploration performance in all the experimental domains, and showing robustness to stochasticity in the dynamics of the environment.

The experiments in low-dimensional continuous-control tasks, where we evaluate the coverage of the environment's state space, have shown that our method provides more in-depth exploration than other methods. LBS provided the most effective and efficient exploration in the Mountain Car and Ant Maze tasks, and strongly outperformed all methods in the more complex HalfCheetah task. Comparing LBS to VIME and Disagreement, we showed that Bayesian surprise in a latent representional space outperforms information gain in parameter space.

In the arcade games results, we showed that LBS works well in high-dimensional settings. By performing best in 5 out of 9 games, compared to several surprisal-based baselines, we demonstrate that the curiosity signal of LBS, based on Bayesian surprise, generally works better than surprisal.   

We also tested LBS to be resilient to stochasticity in the dynamics, both qualitatively and quantitatively. While other methods based on the information gained in parameter space also showed to be robust in the stochastic settings, the exploration performance of LBS are unmatched in both variants of stochastic Mountain Car.  We believe stochasticity is an important limitation that affects several exploration methods and future work should focus on understanding to which extent limitations apply and how to overcome them. 

\clearpage


\clearpage
\newpage 

\section*{Acknowledgements}

This research received funding from the Flemish Government (AI Research Program). Ozan Catal is funded by a Ph.D. grant of the Flanders Research Foundation (FWO).

\bibliography{mybib}

\newpage

\appendix

\section{Zero-shot Adaptation}

In these experiments, adapted from \cite{Sekar2020Plan2Explore}, an exploration policy is trained with intrinsic motivation, using a model-based RL algorithm, based upon Dreamer \cite{Hafner2020Dreamer}. The exploration policy is used to collect experience from the environment, which is in turn used to improve the agent model. Trajectories sampled from the model, including predictions of the environment's rewards, are used to train an exploitative policy that is used sporadically and only for zero-shot evaluation (except Dreamer baseline). 

The model-based RL algorithm trains a variational latent dynamics model of the environment. As a consequence, both the prior and posterior models are already available and we can compute the intrinsic motivation term for LBS in the agent's model directly, with no need to train extra components, in contrast to Plan2Explore that requires training an additional ensemble of predictors.

To compare to the baselines from \cite{Sekar2020Plan2Explore}, we follow the hints from the official Plan2Explore's repository\footnote{\url{https://github.com/ramanans1/plan2explore}} and implement LBS on top of the TensorFlow2 implementation\footnote{\url{https://github.com/danijar/dreamerv2} \cite{Hafner2021DreamerV2}}. For a fair comparison, we replace the settings of the newer repository with the hyperparameters and design choices from the original paper (i.e. hidden size of the GRU of 400, gaussian latent space, latent dimensionality of 60, ...). Additionally, we normalize the LBS intrinsic motivation dividing it by a running mean estimate. 

\begin{figure*}[b!]
    \centering     
    \includegraphics[width=\linewidth]{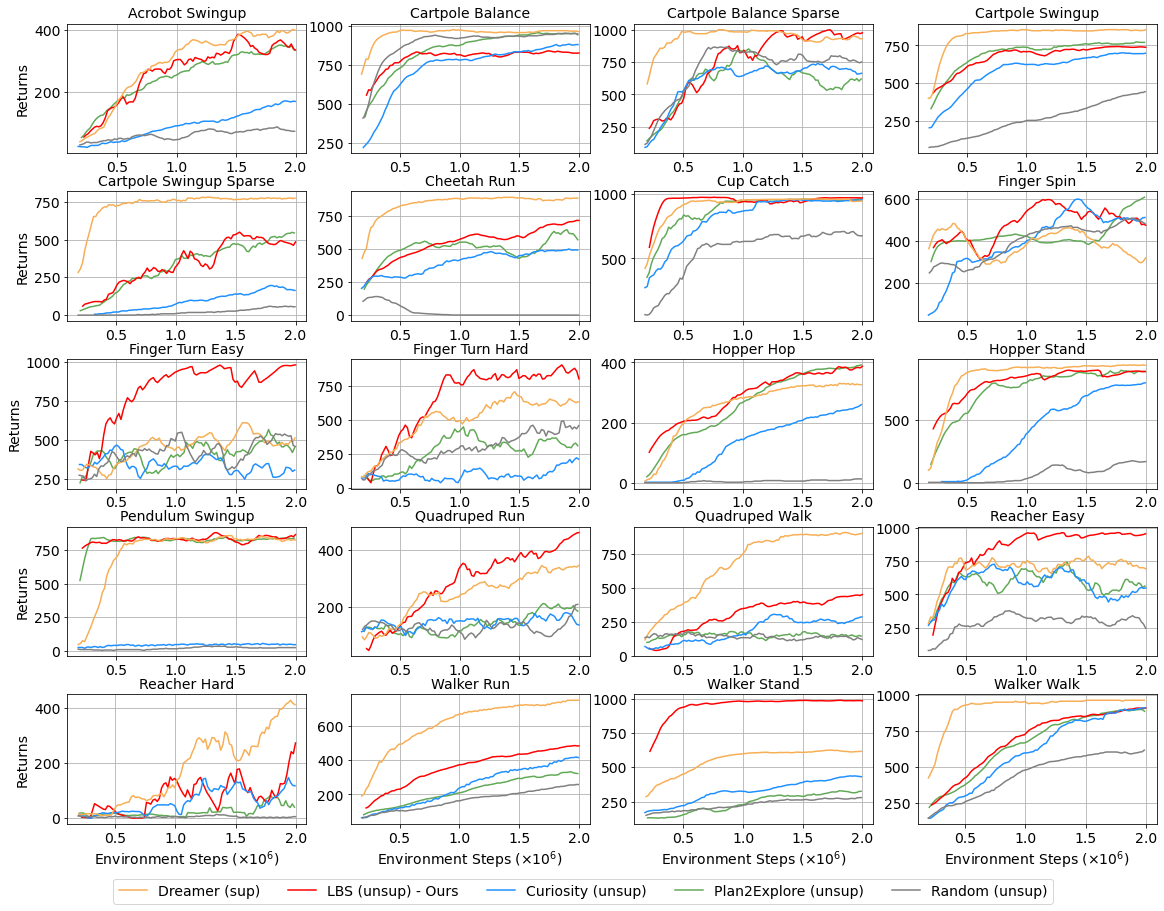}
    \caption{\textbf{Zero-shot experiments on the DMC suite.} Experimental settings and results adapted from \cite{Sekar2020Plan2Explore}.}
    \label{fig:zero_shot}
\end{figure*}

The results are presented in Figure \ref{fig:zero_shot}. Curves for the Dreamer, Curiosity and Random baselines are reverse-engineered from the original paper \cite{Sekar2020Plan2Explore}. In almost every task, using LBS is the best unsupervised exploration strategy. In several tasks, the data collected through LBS-driven exploration allows to outperform  (e.g. Finger Turn) or learn faster (e.g. Pendulum Swingup) than the supervised Dreamer baseline, which has access to the environment rewards also for data collection.



\clearpage

\section{Episodic Counts for Continuous Control}

For the continuous control experiments, we present an additional comparison against the NGU \cite{Badia2020NGU} and RIDE \cite{Raileanu2020RIDE} methods, which, as shown in Table \ref{tab:methods}, also employ episodic counts for their intrinsic reward signals. In continuous state spaces, computing exact state counts is poorly informative, as two states are rarely identical. Thus, we implement episodic counts as pseudo-counts following the procedure presented in NGU for both methods. We employ inverse-dynamics features for computing pseudo-counts for NGU while we directly use the states for RIDE.

\begin{figure*}[t!]
    \centering     
    \includegraphics[width=0.9\linewidth]{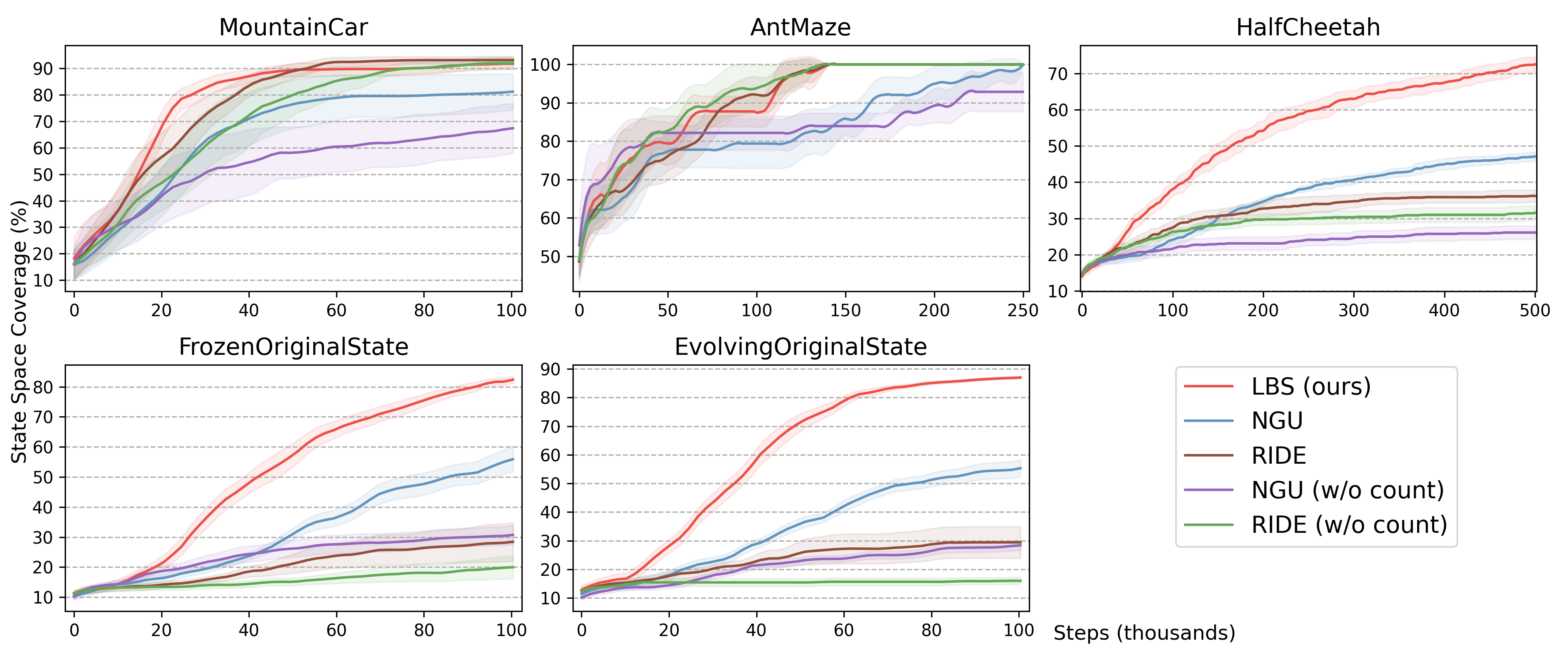}
    \caption{\textbf{Episodic Counts for Continuous Control.} Comparison between LBS, NGU and RIDE, with the latter two being tested both with (original methods) and without (ablations) the episodic-count modulation component.}
    \label{fig:episodic}
\end{figure*}

Results are presented in Figure \ref{fig:episodic}, showing also ablations with respect to the episodic counts modulation (for NGU, we use the results of RND). We see that RIDE performs well in MountainCar and AntMaze but struggles in the higher-dimensional HalfCheetah environment and in the stochastic benchmarks. NGU consistently underperforms LBS but leads to better exploration than RIDE in HalfCheetah and the stochastic benchmarks. Overall, the episodic count component improves performance both for NGU and RIDE across all environments, especially in stochastic settings.   

\section{Pixel-based Models for Arcade Games}
\label{app:pixels}

\begin{figure*}[t!]
    \centering     
    \includegraphics[width=0.9\linewidth]{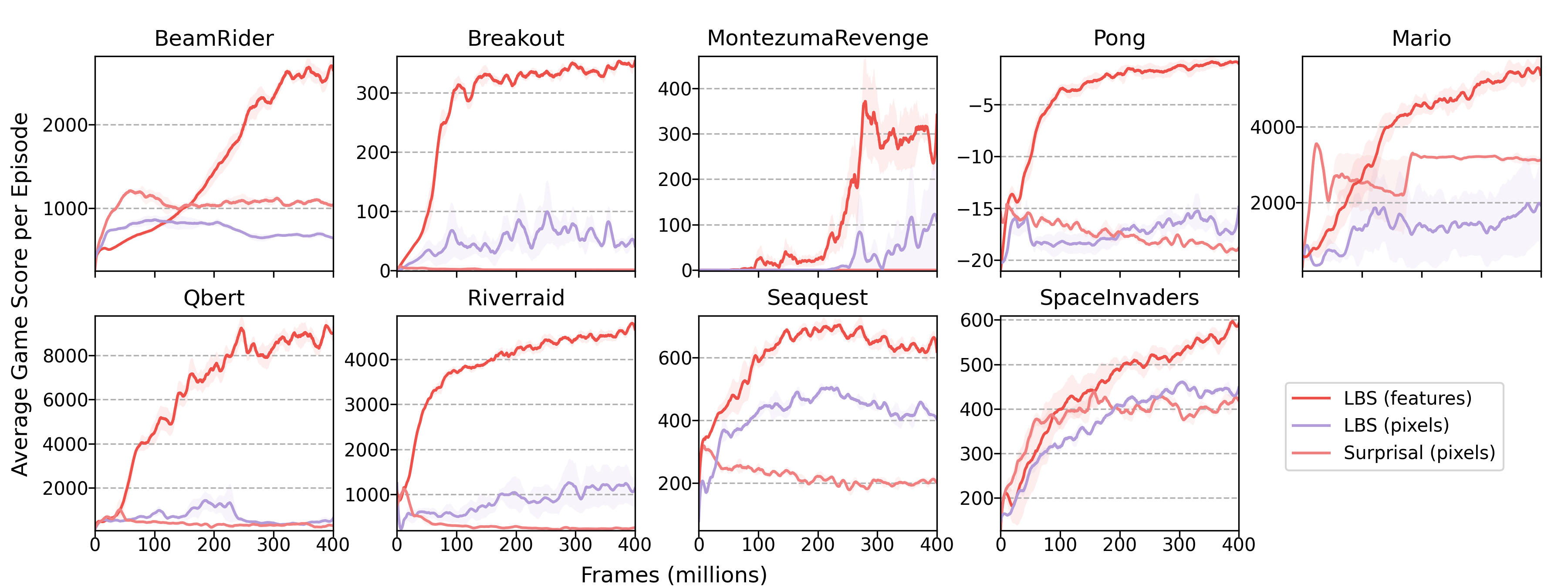}
    \caption{\textbf{Comparison with pixel-based models.} A comparison of pixel-based LBS against feature-based LBS and pixel-based surprisal, on the Arcade games tasks. Lines show the average game score per episode (standard deviations in shade). }
    \label{fig:arcade_pixels}
\end{figure*}

As discussed in the Experiments section, we find that reconstructing features generally works better than reconstructing pixels, for exploration. This finding was also claimed in \citep{Burda2019LSCuriosity}, where they compared different feature-based surprisal methods against pixel-based surprisal.  

In our experiments with pixel-based LBS, we use a "deconvolutional" network as the reconstruction model with an architecture that is the transpose of the posterior's encoder (see later Appendix sections for details). In the LBS loss, we use a value of $\beta=1$. We present results of the pixel-based LBS model in Figure \ref{fig:arcade_pixels}, compared with the LBS feature-based model and the surprisal in pixel space baseline from \citep{Burda2019LSCuriosity}. 

Both pixel-based methods perform significantly worse than LBS in feature space, in all video games. One of the causes may be that is more difficult to learn a dynamics model in pixel space, and errors in the predictions are generally dominated by irrelevant details rather than task-relevant elements. Nonetheless, pixel-based LBS still performs better than pixel-based surprisal in 5 out of 9 tasks (Breakout, Montezuma Revenge, Pong, Riverraid and Seaquest) and comparably in two others (Qbert and Space Invaders). 

We believe that the peformance of pixel-based LBS could further be improved by either significantly improving the reconstruction model in order to provide almost pixel-perfect reconstructions, or by switching to a likelihood free model, e.g. contrastive learning. Investigating these lines of research will be part of future studies.

\section{Latent Dynamics Lower Bound}
\label{app:derivation}
The LBS model's loss is obtained by maximizing a variational lower bound (ELBO) on future states log-likelihood. The derivation employs: marginalization (row 1), the introduction of a variational posterior distribution (row 2), Jensen's inequality (row 3), logarithm properties and the Kullback-Leibler divergence definition (row 4):

\begin{equation*}
    \begin{split}
        & \log p(s_{t+1}|s_t, a_t) \triangleq \log\E_{\rz_{t+1}\sim p(z_{t+1}|s_t, a_t)}[p(s_{t+1}|z_{t+1})] \\
        &= \log \E_{\rz_{t+1}\sim q(z_{t+1})} \left[[p(z_{t+1}|s_t, a_t) p(s_{t+1}|z_{t+1})/ q(z_{t+1})] \right] \\
        &\geq \E_{\rz_{t+1}\sim q(z_{t+1})} \left[\log [p(z_{t+1}|s_t, a_t) p(s_{t+1}|z_{t+1})/ q(z_{t+1})] \right] \\
        &= \E_{\rz_{t+1}\sim q(z_{t+1})} [\log p(s_{t+1}|z_{t+1})] \\ & \quad \quad \quad \quad - \KL [q(z_{t+1}) \Vert p(z_{t+1}|s_t, a_t) ] \\
    \end{split}
\end{equation*}

In the main text, we also assume $q(z_{t+1}|s_t, a_t, s_{t+1})$ to be our approximate posterior. This choice exploits the idea of amortized variational inference, where, instead of optimizing a set of parameters for each state, we introduce a network function that maps states to their latent variable distributions.

\section{Environment Details}
\label{app:env_details}

\textbf{Mountain Car.} The agent is spawned at the bottom of a valley and can move across two hills, by adopting an adequate action. The state space is two-dimensional, and limited to the following ranges: [-1.2, 0.6] for the position and [-0.07, 0.07] for the velocity. We divide each range in ten equally-sized ranges and generate 100 bins from their possible combinations.

\textbf{Ant Maze.} A robotic ant should navigate a `C-shaped' maze, where 7 visitable bins have been identified by heavily discretizing the state space \citep{Shyam2019MAX}.

\textbf{Half-Cheetah.} Tasks using the HalfCheetah environment usually concern training a bipedal walker robot to learn skills such as walking, running, flipping backward or forward. For this reason, we chose to evaluate exploration performance in terms of x-axis velocity and angular velocity, empirically choosing ranges of [-9.0, 9.0] and [-18.0, 18.0], respectively. Each range is divided into ten smaller subranges and exploration is measured as the percentage of the bins visited among the 100 bins, representing all visitable combinations of subranges.

\textbf{Arcade Games.} We follow the setup of \citep{Burda2019LSCuriosity}. Frames from the games are converted to grayscale and resized to 84x84 images. States are obtained by stacking four games frames.

\section{Implementation Details}
\label{app:impl_details}

\textbf{Model Details.} In the distributional layers, standard deviations are obtained by applying a softplus operator to the outputs of the network. The reparametrization trick \cite{Kingma2014VAE} is used to propagate gradients through the sampling from the posterior. 

For vision-based tasks, states are first processed by convolutional networks (CNN) that follow the architecture presented in Table \ref{tab:cnn}. The CNNs are trained as part of the latent prior and posterior models.

Latent prior and posterior are 2-layers MLP networks, while the reconstruction model is a linear layer. For vision-based tasks, we use 512-dimensional hidden layers and latent variables, and LeakyReLU activations. For continuous-control experiments, we use 32-dimensional hidden layers, ReLU activations and we set the latent dimensionality to be the same as the state dimensionality.  
\begin{table}[h]
\begin{center}
\begin{tabular}{|c|c|c|}
\hline \multicolumn{1}{|c}{}  &\multicolumn{1}{|c}{Arcade Games} &\multicolumn{1}{|c|}{MNIST}\\ \hline  
1st layer         & 8x8, 32 ch., stride 4 & 3x3, 32 ch., stride 1\\ \hline
2nd layer         & 4x4, 64 ch., stride 2 & 3x3, 64 ch., stride 1  \\ \hline
3rd layer         & 3x3, 64 ch., stride 1 & maxpool 2x2, stride 1 \\ \hline
\end{tabular}
\end{center}
\caption{Convolutional Network Setup}
\label{tab:cnn}
\end{table}

\textbf{Normalization.} States normalization is applied differently for different kinds of environment. In continuous control tasks, states are normalized by maintaining running statistics. For video games, following the setup in \citep{Burda2019LSCuriosity}, state's mean and standard deviation values are computed on 10,000 steps collected using a random agent, and are kept fixed during training.

Rewards are normalized dividing them by a running estimate of the standard deviation of the returns. We also found beneficial to clip rewards in the range $(-3,3)$ to increase stability.

\textbf{Hyper-parameters.} In all experiments, the same learning rate, number of updates and of mini-batches have been used both for the policy and the model training. 

For continuous-control experiments, we use a learning rate of $3\cdot10^{-4}$, 10 updates per episode, and 32 minibatches. From one environment, we collected $2048$ steps per epoch. For the arcade games experiments, following the setup of \citep{Burda2019LSCuriosity}, we use a learning rate of $1\cdot10^{-4}$, 3 updates per episode, and 8 minibatches. From 128 parallel environments, we collected $128$ steps from each, per epoch. The policy value loss coefficient is fixed to $0.5$ and the entropy coefficient to $0.001$.

We stick as much as possible to standard hyperparameters for all baselines. For Disagreement, we use an ensemble of 5 models. For the LBS' loss we use: $\beta=0.1$ for continuous control experiments, $\beta=0.01$ in the arcade games experiments, and $\beta=2$ in the stochastic tasks. Values of $\beta$ have been searched in the range ($1\cdot10^{-4},10)$, though we did not perform extensive hyperparameters optimization, due to computational constraints.

\section{Additional Experimental Results}
\label{app:experiments}

\begin{figure*}[htb]
    \centering
    \includegraphics[width=.7\textwidth]{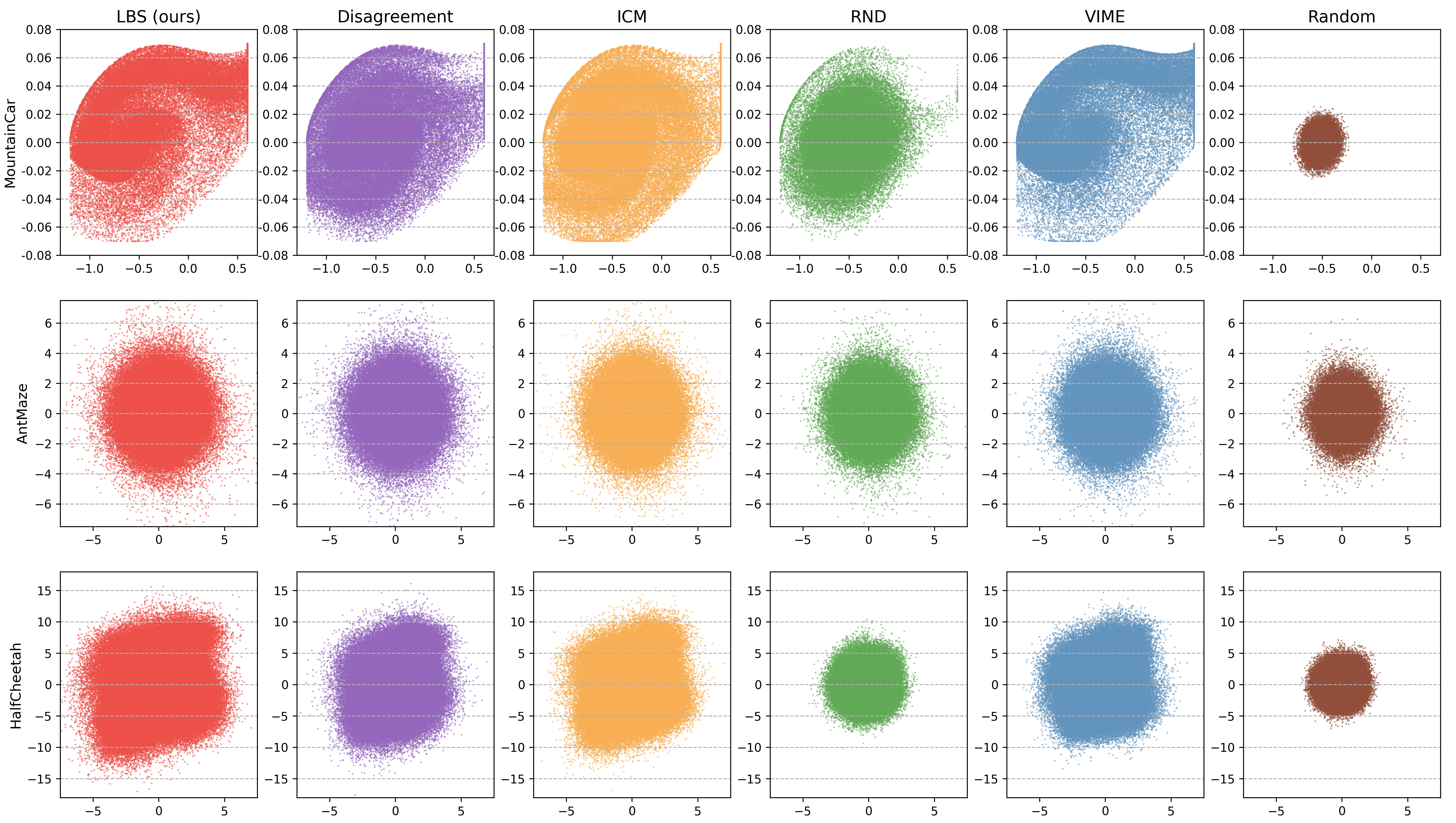}
    \caption{\textbf{Control Plots} On the x-y axis we plot the metrics used to identify the exploration bins of the environment. For AntMaze, these are (x,y) coordinates of the robotic ant. For the other environments, see the Environment Details Appendix.}
    \label{fig:control_plots}
\end{figure*}

In the continuous control experiments, each run was completed within some minutes on a i7-10850H CPU, experiencing similar training times for ICM and RND: around 2.5 minute for MountainCar, 10 minutes for AntMaze, and 14 minutes for HalfCheetah, for each run. LBS usually takes 5-10\% more time, Disagreement is $20$ to $30\%$ slower and VIME is the slowest, being about $200\%$ slower than ICM and RND, in all experiments. 

To provide more insights into the state-space exploration provided by the different methods, we present control plots in Figure \ref{fig:control_plots}. The graphs display state-space visits averaged over the eight random seeds of the experiments, plotted along the two dimensions used to discretize the state space, in Mountain Car and Half Cheetah, and along (x,y) coordinates in the maze, for Ant Maze. 

\begin{figure}[b!]
    \centering
    \includegraphics[width=\columnwidth]{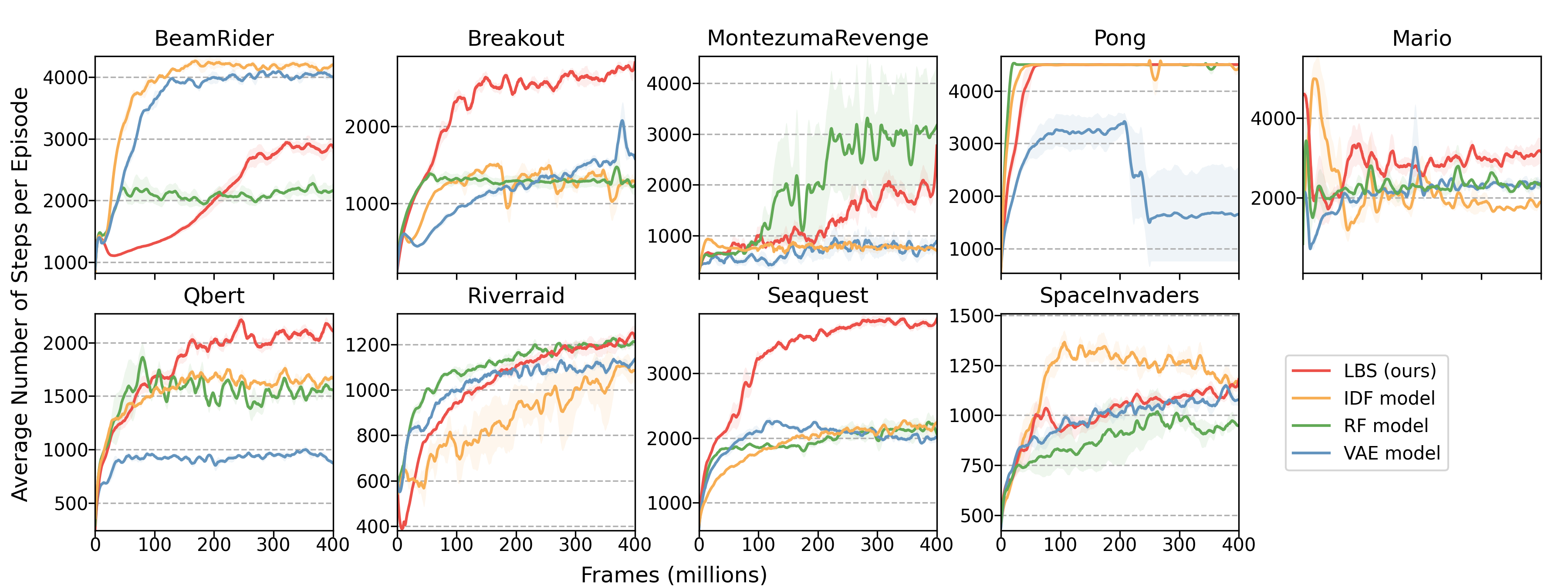}
    \caption{\textbf{Arcade Games}. Average episode lengths during training.}
    \label{fig:arcade_lengths}
    \centering     
    \includegraphics[width=\columnwidth]{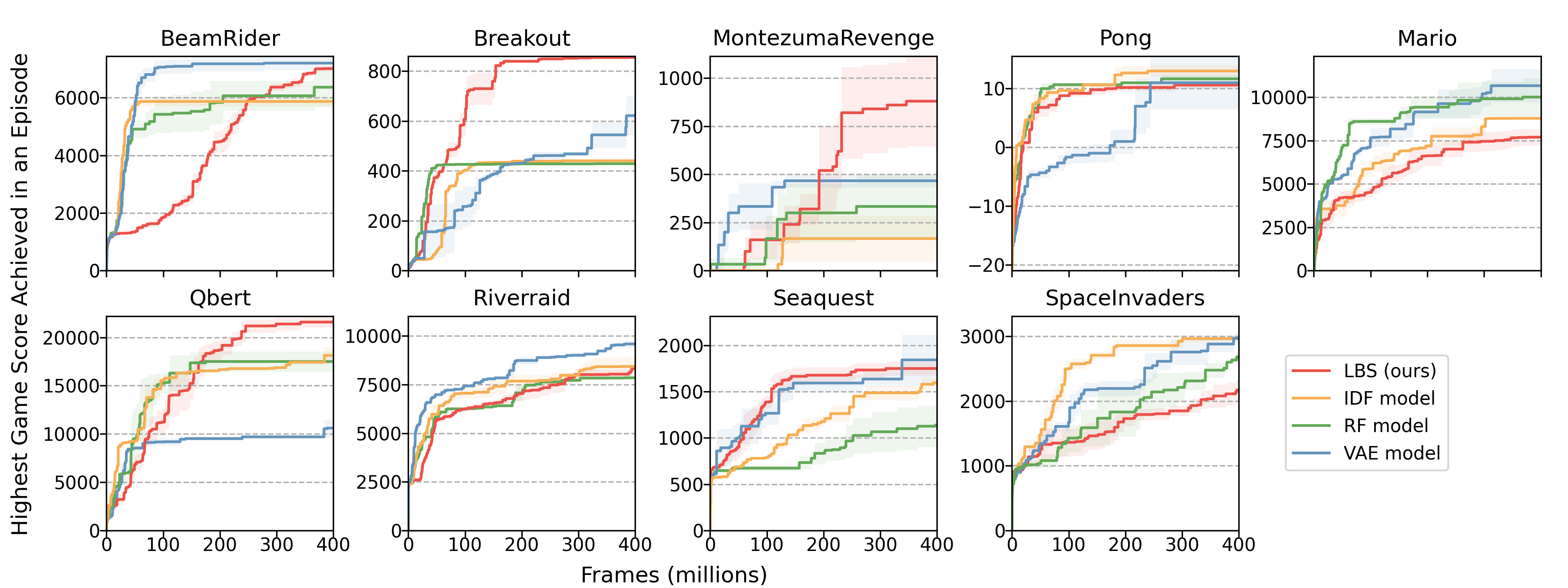}
    \caption{\textbf{Arcade Games}. Average best scores obtained during training.}
    \label{fig:arcade_best_scores}
\end{figure}

For the Arcade Games, we additionally report the curves of the average episode lengths (Figure \ref{fig:arcade_lengths}) and best scores achieved (Figure \ref{fig:arcade_best_scores}), during training. For each experimental run, a training time of around 16 hours on a V100 GPU coupled with a 16-cores CPU was required. 

\end{document}